\documentclass[a4paper,conference]{IEEEtran}
\ifCLASSINFOpdf
\else
\fi
\IEEEoverridecommandlockouts

\usepackage{times}
\usepackage{graphicx}
\usepackage{algorithm}
\usepackage{algpseudocode}
\usepackage{amsfonts}
\usepackage{amsmath}
\usepackage{amssymb}
\usepackage{varwidth}
\usepackage{caption}
\usepackage{color}

\makeatletter
\renewcommand\footnoterule{%
  \kern-3\p@
  \hrule\@width.8\columnwidth
  \kern2.6\p@}
  \makeatother

\begin{document}
%
\title{Generating Image Sequence from Description with LSTM Conditional GAN}

\author{\IEEEauthorblockN{Xu Ouyang\IEEEauthorrefmark{2}, Xi Zhang\IEEEauthorrefmark{2}, Di Ma, Gady Agam}\thanks{asdfasdf}
\IEEEauthorblockA{
Illinois Institute of Technology\\
Chicago, IL 60616\\
\{xouyang3, xzhang22, dma2\}@hawk.iit.edu, agam@iit.edu}
}


%


\twocolumn[{%
\renewcommand\twocolumn[1][]{#1}%
\maketitle
\begin{center}
    \centering
    \includegraphics[width=1.0\textwidth,height=5.8cm]{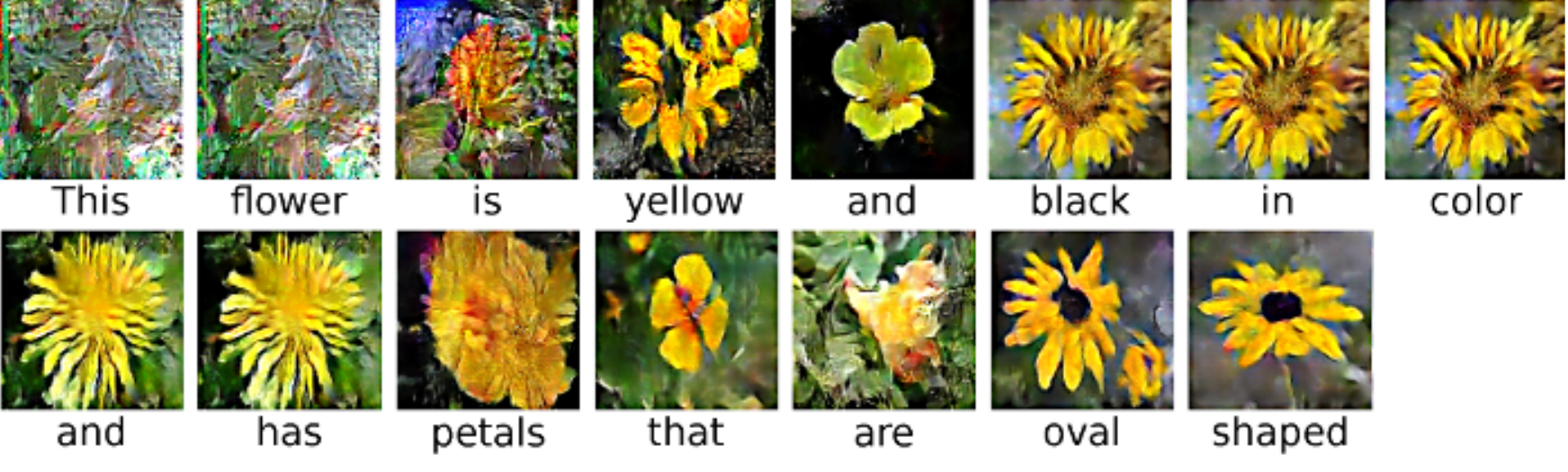}
    \captionof{figure}{Example of image sequence generated according to text description provided to the proposed LSTM conditional GAN. Each image is generated corresponding to  text description up to each word in the sentence.}
\end{center}%
}]

\begin{abstract}
Generating images from word descriptions is a challenging task. Generative adversarial networks(GANs) are shown to be able to generate realistic images of real-life objects. In this paper, we propose a new neural network architecture of LSTM Conditional Generative Adversarial Networks to generate images of real-life objects. Our proposed model is trained on the Oxford-102 Flowers and Caltech-UCSD Birds-200-2011 datasets. We demonstrate that our proposed model produces the better results surpassing other state-of-art approaches.
\end{abstract}

\let\thefootnote\relax\footnotetext{\IEEEauthorrefmark{2}Denotes first two authors have equal contribution.}


%
\IEEEpeerreviewmaketitle

\section{Introduction} 
Generating realistic images from word descriptions is a fundamental problem with useful applications in image reconstruction, image search, portrait drawing and so on. For example, given the description: ``this flower has flat long and skinny yellow petals in one slightly extended ring configuration", we can generate a corresponding realistic image shown in Figure 1. Deep neural networks such as the Generative Adversarial Networks (GANs) \cite{goodfellow2014generative}, have been shown to be able to generate realistic images from their corresponding descriptions.

  Previous approaches of image generation have various problems in image generation. For instance, it is hard to create high resolution and realistic images from descriptions; since natural image distribution and potential model distribution may not overlap in high dimensional pixel space. While some neural network such as \cite{reed2016generative} by Reed et al. can generate realistic images, they need train multiple models to synthesize images. The proposed approach generates images gradually following the evolving structure of the sentence.

\subsection{Motivation}   
We use the LSTM network in the proposed approach to extract information from each word of description following the sequence in the description. We train a conditional GAN network to synthesize the image using the description features learned up to each word of the description. 

\begin{figure}[h]
\centerline{
\begin{tabular}{c}
  \resizebox{0.48\textwidth}{!}{\rotatebox{0}{
  \includegraphics{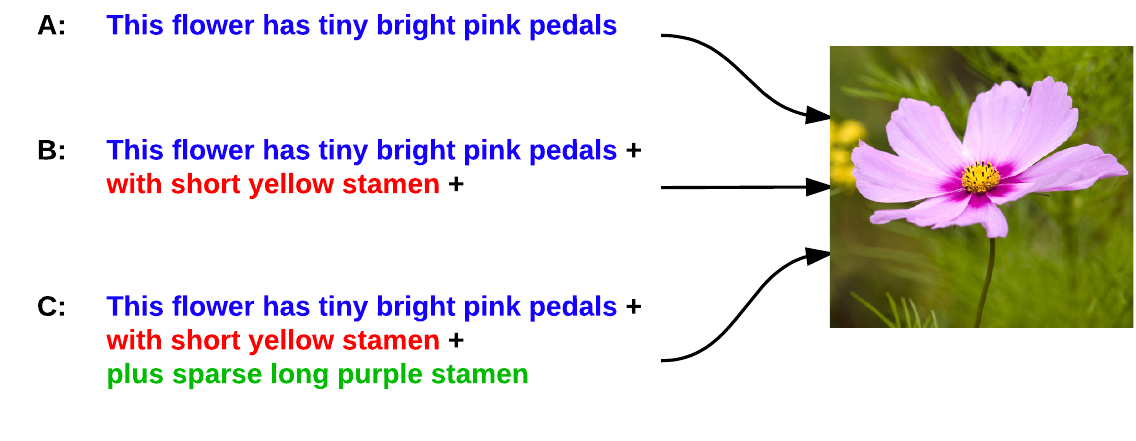}}}
\end{tabular}}
\caption{An illustration of multiple levels of semantics are contained in a given sentence. All of these could possibly correspond to one image.}
\label{fig: sentence mapping}
\end{figure} 

\begin{figure}[h]
\centerline{
\begin{tabular}{c}
  \resizebox{0.48\textwidth}{!}{\rotatebox{0}{
  \includegraphics{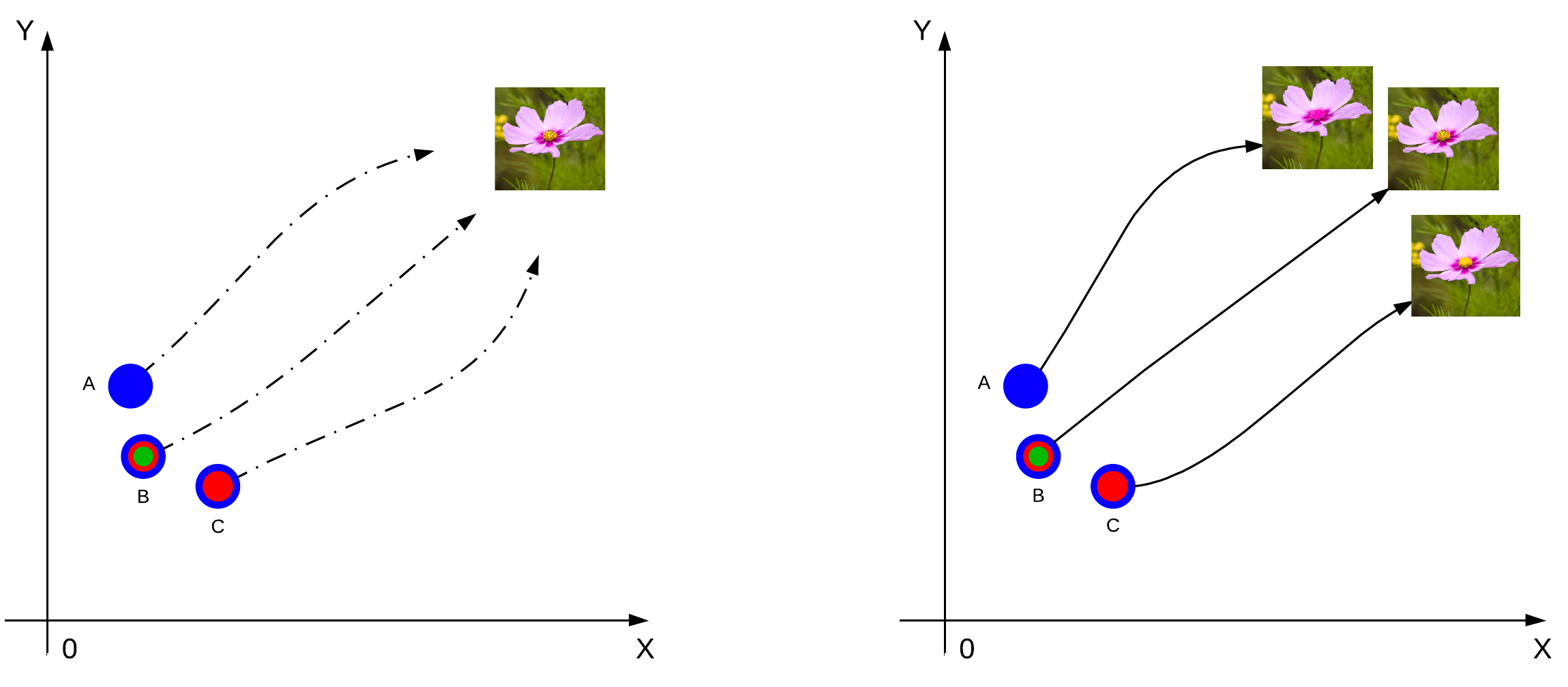}}}
\end{tabular}}
\caption{An illustration of mapping sentences shown in Figure \ref{fig: sentence mapping} to generated images. Colors of circles corresponds to colors of sentences used in Figure \ref{fig: sentence mapping}.}
\label{fig: semantic mapping}
\end{figure} 

A possible question is whether it is reasonable to use the same real image as a target for synthesized images at multiple words. Based on experimental results, we think it is not only possible to do that, but also has several advantages by doing so. LSTM is capable of learning the semantic meaning of sentences in terms of understanding syntactic structures of sentences. While a sentence may contain several sub-sentences which themselves represent complete concepts. Each of these sub-sentences only provides complementary information to other sub-sentences. Therefore, using the identical real image as the target of image synthesis supports a common goal.

An example is given in Figure~\ref{fig: sentence mapping}. In this case, sentence (A) describes basic properties of a flower, where sentences (B) and (C) add additional information gradually. All three sentences can be mapped to a single flower picture on the right. To better understand the mapping, we show a 2D illustration in Figure~\ref{fig: semantic mapping}. In this figure, we use three circles with colors to represent sentences in Figure~\ref{fig: sentence mapping}, and a thumbnail of the flower image to represent the target image in this circumstance. By using the same image as a target for three sentences, GAN maps three colored circles to the image instead of mapping to three images with slight differences. These differences are due to additional information carried by the red sentence and the green sentence as shown in Figure~\ref{fig: sentence mapping}. Although the theoretical explanation of this behavior is still under investigation, we observe this behavior in experimental results of the proposed network. Many results show that sub-sentences are precisely mapped into images with corresponding descriptions of sub-sentences.

Using adversarial loss at each word allows the proposed network to learn additional information introduced by sub-sentences. This special design can produce better details for synthesized image generation. Another advantage of this design is that by computing adversarial loss at each word, we strengthen the gradient flow at each word when the flow back propagates. Both the generator and discriminator get more opportunities to be trained. 
  
\subsection{Related Work} 
\noindent\textbf{LSTM} networks are commonly used in natural language processing. The initial version of the LSTM block~\cite{hochreiter1997long} includes cells, input gate, output gate, and forget gate. The input gate controls the flow of input activation into the memory cell. The output gate controls the output flow of cell activation into the rest of the network. The forget gate was to process continuous input streams that are not segmented into subsequences. Training process is a combination of Real Time Recurrent Learning (RTRL)~\cite{robinson:utility} and Backpropagation Through Time (BPTT)~\cite{werbos1988generalization}. Many variants of LSTM were proposed recently. ~\cite{schmidhuber2007training}\cite{bayer2009evolving}\cite{jozefowicz2015empirical}\cite{otte2014dynamic}
\newline
\newline
\noindent\textbf{GAN} is a popular and successful deep neural network for image generation. This model consists of a generator that generates images from a uniformly distributed random noise, and a discriminator that discriminates between generated images and real images. Even though this model can achieve some realistic images, it is difficult to train. Deep Convolutional GANs (DCGANs) \cite{radford2015unsupervised} implemented an efficient and stable architecture to get striking image synthesis results. Conditional GANs \cite{DBLP:journals/corr/MirzaO14} extended the GANs to a conditional model which makes it possible to direct the data generation process. Stack GANs \cite{DBLP:journals/corr/ZhangXLZHWM16} stacked several GANs for text-to-image synthesis and used different GANs to generate images of different sizes. Some more recent uses of GANs include reference\cite{pathak2016context}\cite{DBLP:journals/corr/MakhzaniSJG15}\cite{arjovsky2017wasserstein}\cite{isola2016image}
\newline
\newline
\noindent\textbf{Data synthesis} Several approaches explored the ability of deep neural network to synthesize high-quality, realistic images. Zhang et al proposed a multi-channel auto-encoder to transfer synthetic data to real data \cite{7424358}.  A general method for using synthetic data features to solve imbalanced learning was introduced in~\cite{Zhang:2016:CCG:2983323.2983789}. \cite{reed2016generative} used text descriptions as a condition instead of class labels based on the conditional GAN, and successfully generated realistic flower and bird images that correspond to their descriptions. Furthermore, they introduced a manifold interpolation regularizer for the GAN generator that significantly improves the quality of the generated samples. However, they need to train a hybrid of a character-level ConvNet with a recurrent neural network(char-CNN-RNN) to get text features as described in \cite{reed2016generative} before image generation network training. In other words, they use two individual steps to achieve image generation which takes too many processes.
  
\subsection{Novel Contribution}
The novel contribution of this work is two-fold. First, we propose a new network by combining the network structures of  LSTM and GAN to convert a sentence description to images. The proposed network structure synthesizes an image for every substructure of a sentence, so that the proposed network can capture critical concepts of a sentence better and result in synthetic images containing better visual details. Both qualitative and quantitative evaluations demonstrate this conclusion. Second, by running the proposed network, synthetic images will be produced for each word in a sentence, which in turn provides visualization of the semantics of the sentence. By analyzing the visual relationship among synthesized images, the visualization offers a valuable way to understand how LSTM parses each word to decode semantics of sentence.

\section{Method}
\begin{figure}[h]
\centerline{
\begin{tabular}{c}
  \resizebox{0.45\textwidth}{!}{\rotatebox{0}{
  \includegraphics{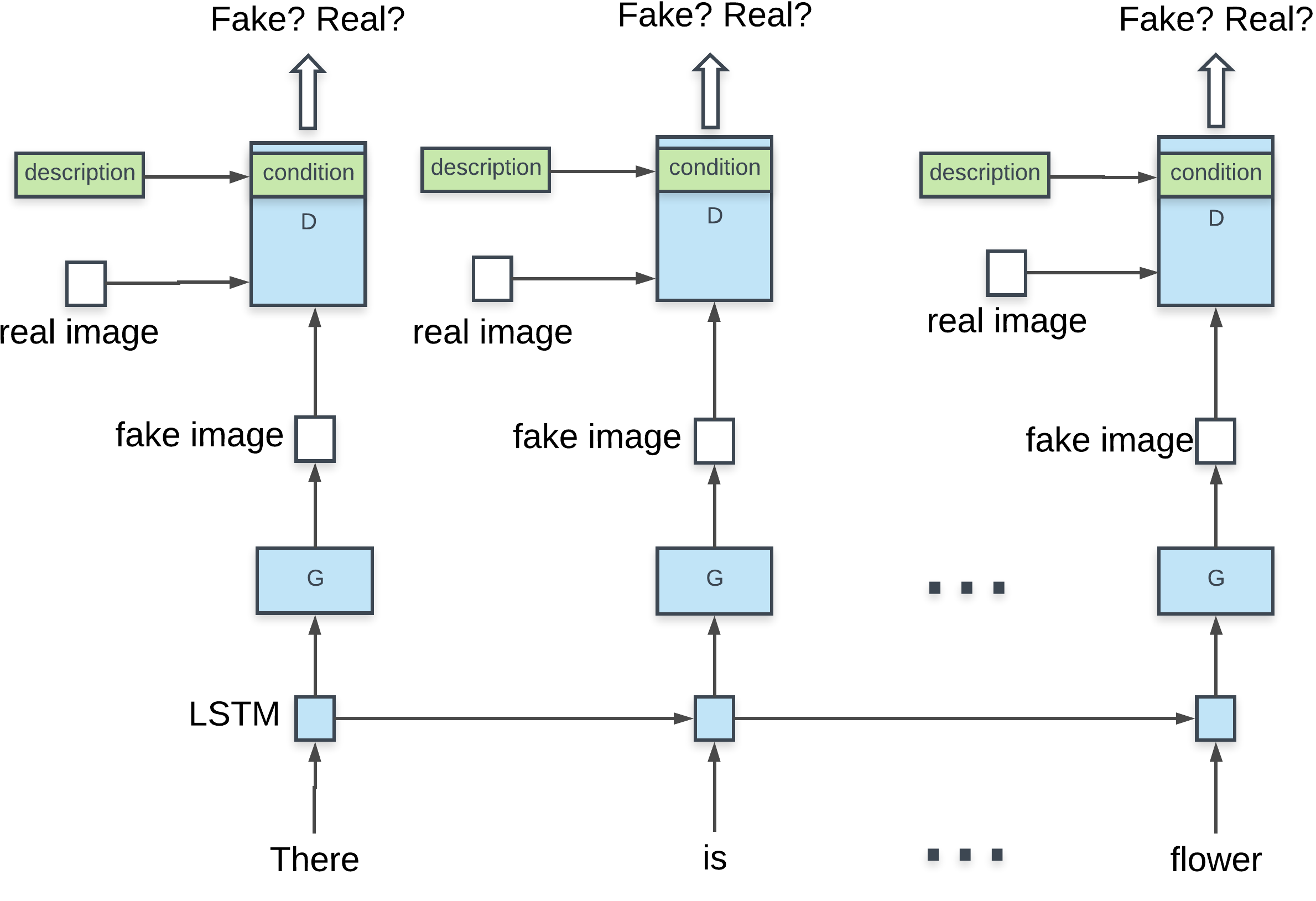}}}
\end{tabular}}
\caption{The proposed LSTM Conditional GANs architecture.}
\label{fig: network architecture}
\end{figure}

The structure of the proposed neural network contains three parts: an LSTM, a generator ($G$) and discriminator ($D$).
We illustrate the architecture of the proposed neural network in Figure~\ref{fig: network architecture}. Instead of using uniform random noise as input, an embedding generated using word2vec \cite{mikolov2013distributed} is computed for each word. The input feature vector of $t$-th word of an input sentence is denoted as $x(t)$. Following \cite{reed2016generative}, a skip-though vector representing the semantic meaning of an entire sentence used as a condition in our work. We denote it as $y\sim p(d)$ where d is the description of an image and y is the skip thought vectors of a description.

\subsection{LSTM}
Information transfer between words is achieved by LSTM which learns semantic meaning and relationships between words in a sentence. Qualitative results in the experimented section show how the learned semantics enables generating the final image gradually word by word. Quantitatively, we show that with this process, the generated synthetic images simulate actual images better than the compared methods. In the proposed network architecture, LSTM is connected with input feature vectors and followed by a generator. Connecting input feature vectors directly enables LSTM to capture essential semantic information of word features. \cite{hochreiter1997long} Using LSTM before the generator in the network structure enables learned semantics to correlate more closely with high-level visual features of images, which gives the generator more flexibility to add details to improve the visual quality of the synthesized images. It could be seen from the table that images generated by the proposed network are more similar to the real images which means the proposed network can synthesize images that capture the better content of real images both in large and small scale. 
%
 
Given an input feature of the $t$-th word represented as $x(t)$, we use the formulation of traditional LSTM with input $i_t$, forget $f_t$, and output gates $o_t$. The computation of these three gates are given in \cite{gers1999learning}.


\subsection{GAN}
\noindent\textbf{Generator} A deconvolution network denoted by $G: \mathbb{R}^Z\rightarrow \mathbb{R}^S$. Using $h_t\in \mathbb{R}^Z$ to represent features fed to the generator, we denote images synthesized by the generator as $G(h_t)\in \mathbb{R}^S$.  Instead of using uniformly random noise as input which is then concatenated with text features in a certain layer of the generator, in each timestamp, we feed the generator the word2vec feature of each word from the LSTM. This is done so that the image of each word is generated and fed to the discriminator individually, which can reinforce generator's generation ability. Furthermore, this allows us to observe how image is generated step by step.
\newline
\newline
\noindent\textbf{Discriminator}  Used to distinguish between real and synthesized images. The proposed network generates synthetic images matching the description of input sentences. GAN \cite{goodfellow2014generative} is initially designed as an unconditional generative model, which does not necessarily produce corresponding results matching the input. Mirza and Osindero\cite{DBLP:journals/corr/MirzaO14} proposed an architecture which turns the GAN model to a conditional GAN. In conditional GAN, a corresponding condition is concatenated to hidden features of the discriminator.  We use skip-thought \cite{kiros2015skip} vectors which learned whole sentence embeddings as the corresponding condition in the proposed network. The skip thought vector is generated from an encoder-decoder model that tries to reconstruct the surrounding sentences of an encoded passage. Sentences that share semantic and syntactic properties are thus mapped to similar vector representations. We denote the skip-thought vector as $y\in \mathbb{R}^T$. Given a generated image $G(h_t)$ and a real image labeled as $r$, we represent the outputs of the discriminator using these two inputs under the condition as $D(G(h_t)|y): \mathbb{R}^S \times \mathbb{R}^T \rightarrow \{0, 1\}$ and $D(r | y): \mathbb{R}^S \times \mathbb{R}^T \rightarrow \{0, 1\}$. 

\subsection{Training of the network}
Given the word2vec feature $x_t$ in the LSTM and the skip-thought vector in the discriminator $D$, we train $D$ to maximize the probability of assigning the correct label to both training examples and samples from $G$.  We simultaneously train $G$ to minimize $log(1 - D(G(h)|y))$. In other words, $D$ and $G$ play the following two-player min-max game with value function $V(G, D)$:
\begin{equation}
\begin{split}
\min_{G} \max_{D} V(D, G) = &\mathbb{E}_{r}[log(D(r|y))] + \\ 
&\mathbb{E}_{h_t}[log(1-D(G(h_t)|y))]
\end{split}
\end{equation} 
Where t = 1,...,N, and N is the length of description.

\begin{figure*}[th]
\centerline{
\begin{tabular}{c}
  \resizebox{0.9\textwidth}{!}{\rotatebox{0}{
  \includegraphics{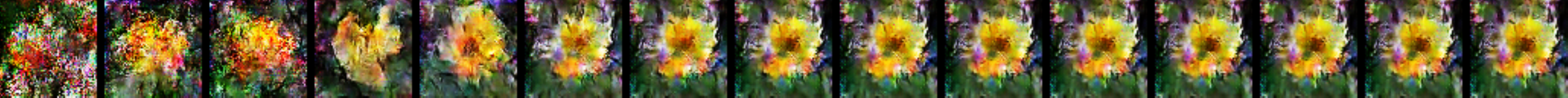}}}
  \\ 
  one yellow flower with multiple layers of yellow petals surrounding one red bundle of stamen
  \\
  \resizebox{0.9\textwidth}{!}{\rotatebox{0}{
  \includegraphics{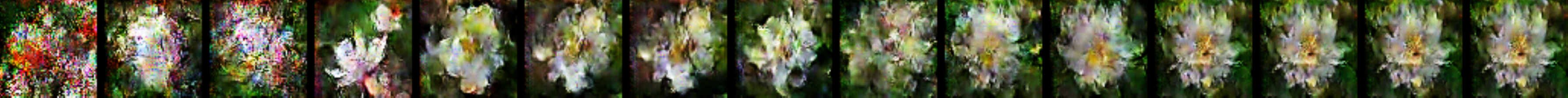}}}
  \\ 
  one white flower with multiple layers of white petals surrounding one red bundle of stamen
  \\
  \resizebox{0.9\textwidth}{!}{\rotatebox{0}{
  \includegraphics{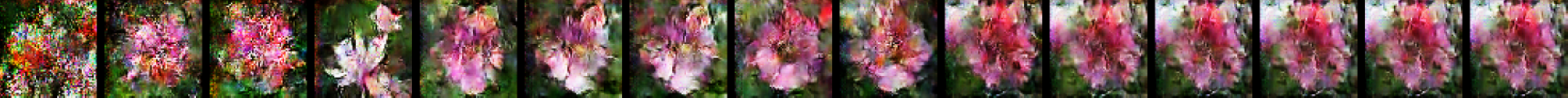}}}
  \\ 
  one pink flower with multiple layers of pink petals surrounding one red bundle of stamen
  \\
\end{tabular}}
\caption{Examples of synthesized images using description with same sentence structure but different colors in description.}
\label{fig: petal color changes}
\end{figure*} 

In \cite{reed2016generative}, to separate the case between \textit{(real image, right text)} and \textit{(real image, wrong text)} as input to the discriminator, a third term is added which penalizes cases of real images with wrong text in the objective function. This additional term is shown to be effective in generating sharper and more accurate image content. Similarly, we employ the same idea in our work. Instead of using the combination of \textit{(real image, wrong text)} as the term, we slightly change it to \textit{(real mismatched image, right text)} and label it as $r^*$ in order to provide the discriminator more opportunities to learn from different images. Thus, the modified objective is:
\begin{equation}
\label{eqn: objective}
\begin{split}
\min_{G} \max_{D} V(D, G) = &\mathbb{E}_{r}[log(D(r|y))] + \\ 
&\mathbb{E}_{h_t}[log(1-D(G(h_t)|y))]  + \\
&\mathbb{E}_{r^*}[{log(1-D(r^*|y))}]
\end{split}
\end{equation} 

At each time step, one word is entered into the LSTM unit. One synthetic image is then generated using the generator, which is immediately fed to the discriminator to compute an adversarial loss for each word. This way of computing adversarial loss for each word in a sentence is a primary contribution of this paper. An alternative way is to only generate one synthetic image at the last word of a sentence and compute adversarial loss just once. Experimental results show that the proposed approach generally synthesizes better quality images than existing methods. 

\begin{figure*}[t]
\centerline{
\begin{tabular}{cc}
  \resizebox{0.9\textwidth}{!}{\rotatebox{0}{
  \includegraphics{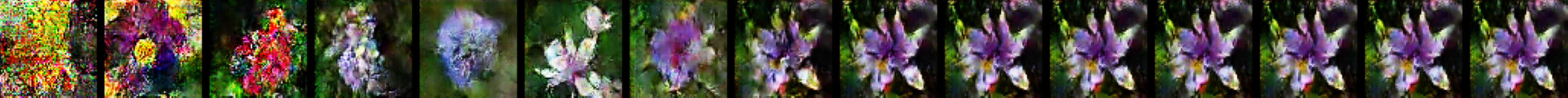}}}
  &
  \resizebox{0.059\textwidth}{!}{\rotatebox{0}{
  \includegraphics{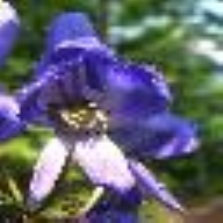}}}
  \\
  this is one purple flower with asymmetrical petals and one tangled yellow and black center &
  \\
  \resizebox{0.9\textwidth}{!}{\rotatebox{0}{
  \includegraphics{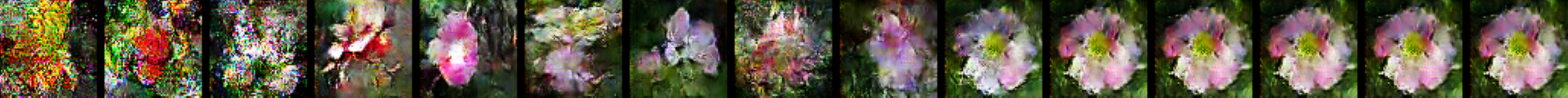}}}
  &
  \resizebox{0.059\textwidth}{!}{\rotatebox{0}{
  \includegraphics{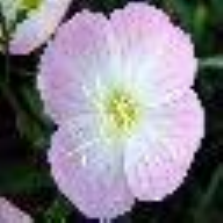}}}
  \\
  this flower has very light petals with pink veins white stamen and one yellow ovary &
  \\
  \resizebox{0.9\textwidth}{!}{\rotatebox{0}{
  \includegraphics{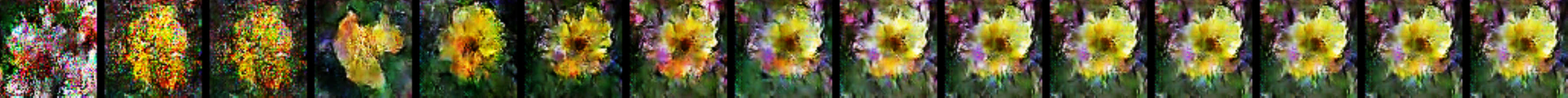}}}
  &
  \resizebox{0.059\textwidth}{!}{\rotatebox{0}{
  \includegraphics{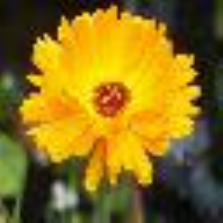}}}
  \\
  one yellow flower with multiple layers of yellow petals surrounding one red bundle of stamen &
  \\
\end{tabular}}
\caption{Examples of generated images using the proposed network. Images at the right most column are real images. }
\label{fig: more results}
\end{figure*} 

The procedure of training the proposed LSTM based text to image synthesis network is presented in Algorithm 1. For a sentence with $n$ words, the algorithm computes a cost shown Equation~\ref{eqn: objective} for each word. The total cost of the entire sentence is a sum of costs from all words. And this is the major contribution of this work. 

\begin{algorithm}
  \caption{Minibatch stochastic gradient descent training of LSTM Conditional GAN}
  \begin{itemize}
    \setlength\itemsep{0.002em}
	\item Given $\theta_l$, $\theta_g$, $\theta_d$ the parameters of LSTM, generator and discriminator.
	\item Given $m$ samples in a batch.
	\item Given $n$ timestamps used in LSTM training.    
  \end{itemize}
  \begin{algorithmic}[1]
    \For{\texttt{number of training iterations}}
    	   \For{\texttt{k steps}}
           \State 
           Update the discriminator  by ascending its stochastic gradient:
           \State
           $\bigtriangledown_{\theta_d}\frac{1}{n}\sum_{t=1}^{n}\frac{1}{m}\sum_{i=1}^m[logD(r^{(i)}|y) + log(1 - D(G(h_t^{(i)})|y)] + log(1 - D(G({r^*}^{(i)})|y))$
           \EndFor
          \State
          Update the generator and LSTM by descending their stochastic gradients respectively:
          \State
		  $\bigtriangledown_{\theta_l}\frac{1}{n}\sum_{t=1}^{n}\frac{1}{m}\sum_{i=1}^mlog(1 - D(G(h_t^{(i)})|y))$
		  \State
		  $\bigtriangledown_{\theta_g}\frac{1}{n}\sum_{t=1}^{n}\frac{1}{m}\sum_{i=1}^mlog(1 - D(G(h_t^{(i)})|y))$
    \EndFor
  \end{algorithmic}
\end{algorithm}

In Algorithm 1, in order to prevent saturation, we update the discriminator two times in every iteration. The reason is that at the beginning of training when the generator is not well trained, the discriminator can reject samples from the generator with high confidence because they are clearly different from the real samples. More experimental results are shown in Figure~\ref{fig: more results}.  

\section{Experiments}
We evaluated the proposed network on the Oxford-102 flower image dataset~\cite{Nilsback08} and Caltech-UCSD birds-200-2011 dataset~\cite{WahCUB_200_2011}. The Oxford-102 dataset contains 8,189 images of flowers from 102 different categories. The Caltech-UCSD Birds-200-2011 has 11,788 bird images in total which is divided into 200 categories. For both datasets, each image has 10 descriptions.

We first trained the word2vec model on a Wikipedia dataset. We then used the pretrained skip-thought vector model on descriptions to get sentence representations which are used as conditions in the discriminator. 

The image size in our experiments is $64\times64\times3$. The learning rate is set to 0.0002. We used the Adam~\cite{kingma2014adam} solver with momentum 0.5 for back propagation. The network is trained 600 epochs with each minibatch size set as 64.

\subsection{Analysis of generated images}
The primary goal of this paper is to learn a model that can understand the semantic meaning contained in each sub-structures of a sentence, to support an ability to synthesize realistic images. The proposed network can show visualization results for a sentence word by word, which is the unique contribution of our work. Two experiments were conducted to verify the proposed model.

\begin{figure}[h]
\centerline{
\begin{tabular}{cc}
  \resizebox{0.23\textwidth}{!}{\rotatebox{0}{
  \fbox{\includegraphics{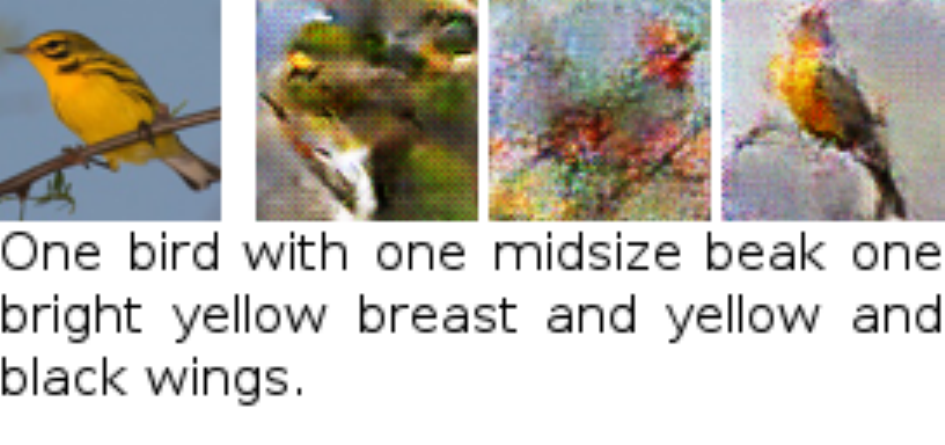}}}}
  &
  \resizebox{0.23\textwidth}{!}{\rotatebox{0}{
  \fbox{\includegraphics{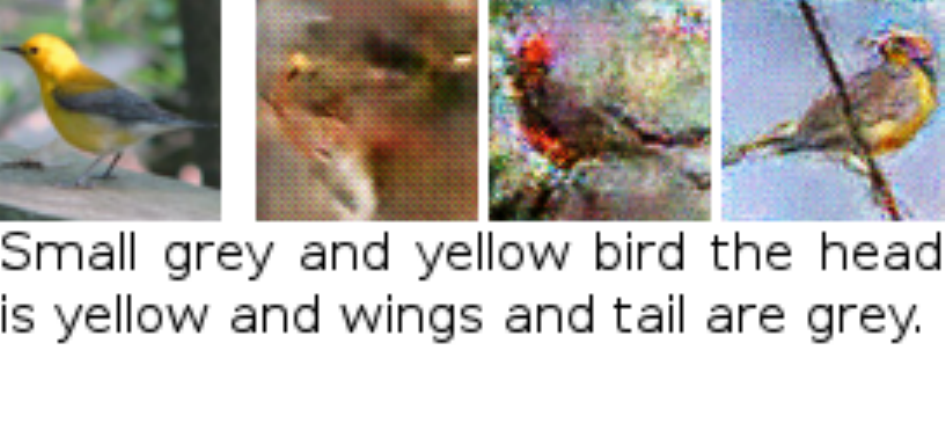}}}}
  \\
  \resizebox{0.23\textwidth}{!}{\rotatebox{0}{
  \fbox{\includegraphics{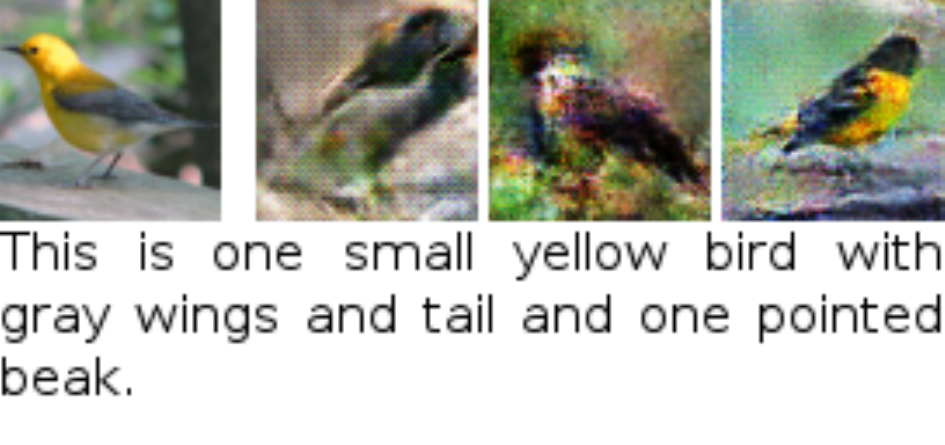}}}}
  &
  \resizebox{0.23\textwidth}{!}{\rotatebox{0}{
  \fbox{\includegraphics{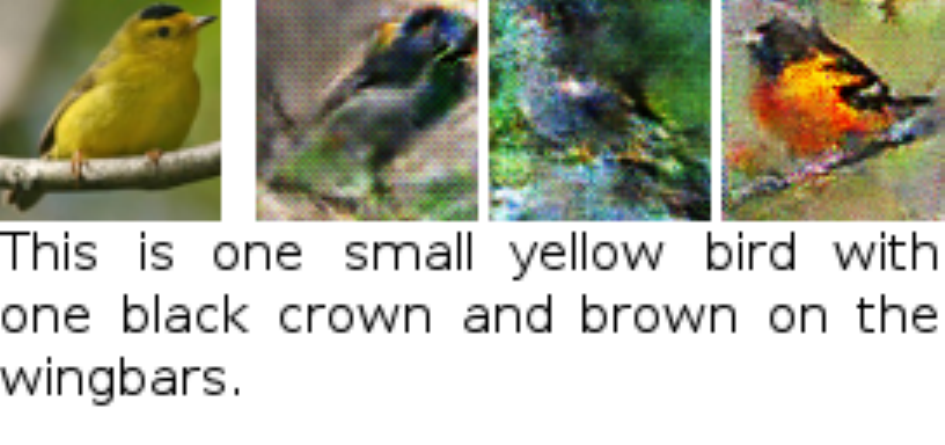}}}}	
  \\
  \resizebox{0.23\textwidth}{!}{\rotatebox{0}{
  \fbox{\includegraphics{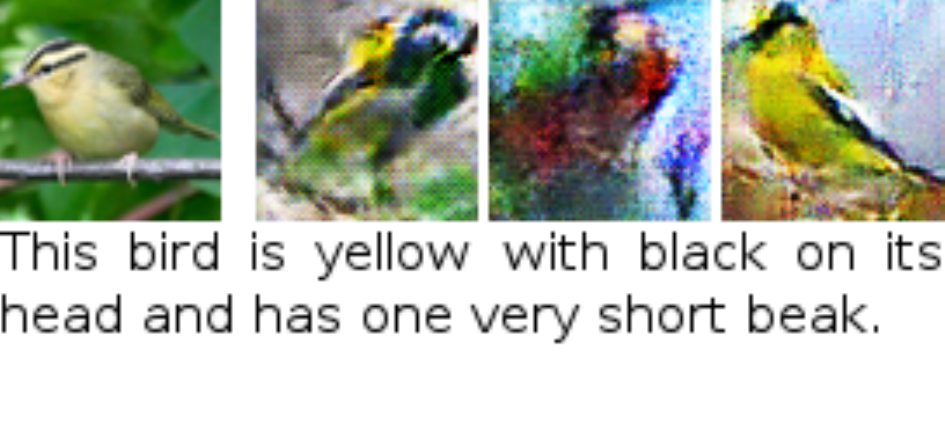}}}}
  &
  \resizebox{0.23\textwidth}{!}{\rotatebox{0}{
  \fbox{\includegraphics{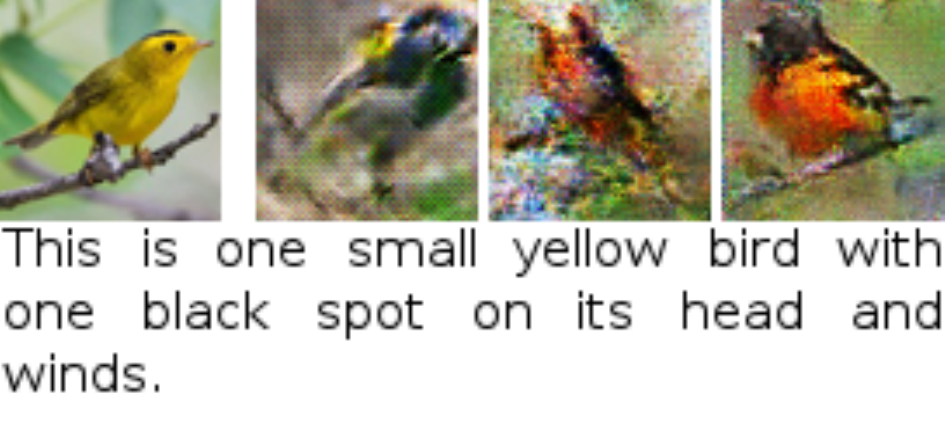}}}}	
  \\
  \resizebox{0.23\textwidth}{!}{\rotatebox{0}{
  \fbox{\includegraphics{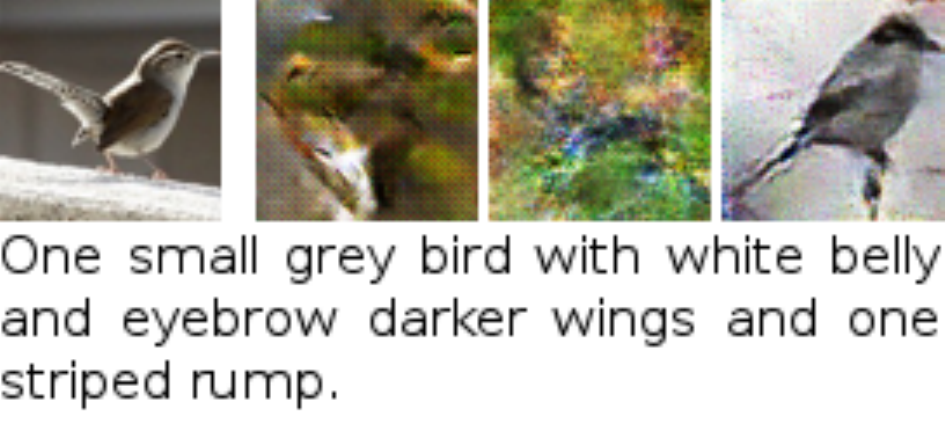}}}}
  &
  \resizebox{0.23\textwidth}{!}{\rotatebox{0}{
  \fbox{\includegraphics{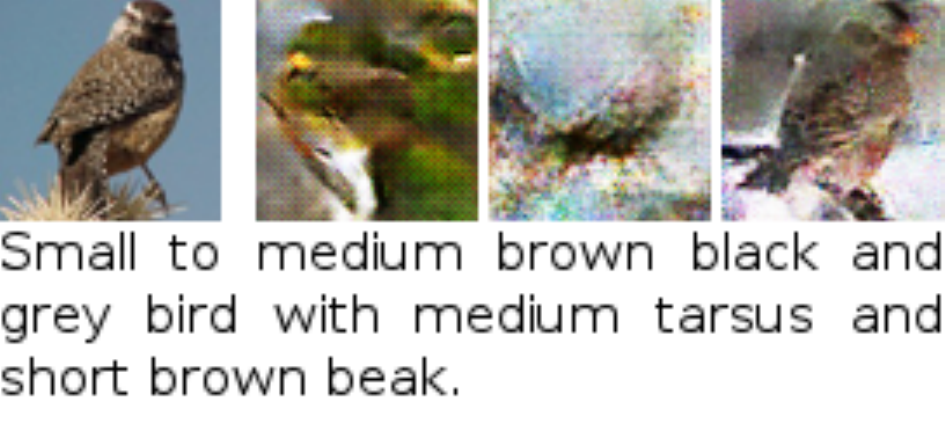}}}}	
  \\
\end{tabular}}
\caption{Comparison with \cite{reed2016generative} and generating image using output of the last step of the proposed LSTM conditional GAN (labeled as last-word). In each group from left to right: true image, \cite{reed2016generative}, last-word, and the proposed approach.}
\label{fig: cgan comparison}
\end{figure}

First, to see whether the proposed network can understand the semantic meaning expressed in sentences, we generated images with the identical structure of sentence but three different colors of petal: yellow, white and pink. In Figure~\ref{fig: petal color changes}, we can observe that the petal color of synthesized flowers all correspond to their descriptions.

Next, in order to examine whether the network can generate image gradually, we use different structure and meaning of descriptions. In Figure~\ref{fig: more results}, we generated images word by word. As we progress in a sentence, we observe that the generated image looks more and more like the description.
  
\subsection{Qualitative comparison}
The two most well-known works relevant to ours are Conditional GAN ~\cite{reed2016generative} and Conditional DRAW Network ~\cite{mansimov2015generating}. Since the results generated by the DRAW network only represent a rough concept of a given description, the synthesized image is not realistic. Thus we compare the proposed network to a conditional GAN in this work. Also, it is truly beneficial to generating images at each word of a sentence using the proposed LSTM conditional GAN, in Figure \ref{fig: cgan comparison} we also compared to images generated up to the last step of the proposed approach (labeled as Last-word)

\begin{figure}[th]
\centerline{
\begin{tabular}{cc}
  \resizebox{0.24\textwidth}{!}{\rotatebox{0}{
  \includegraphics{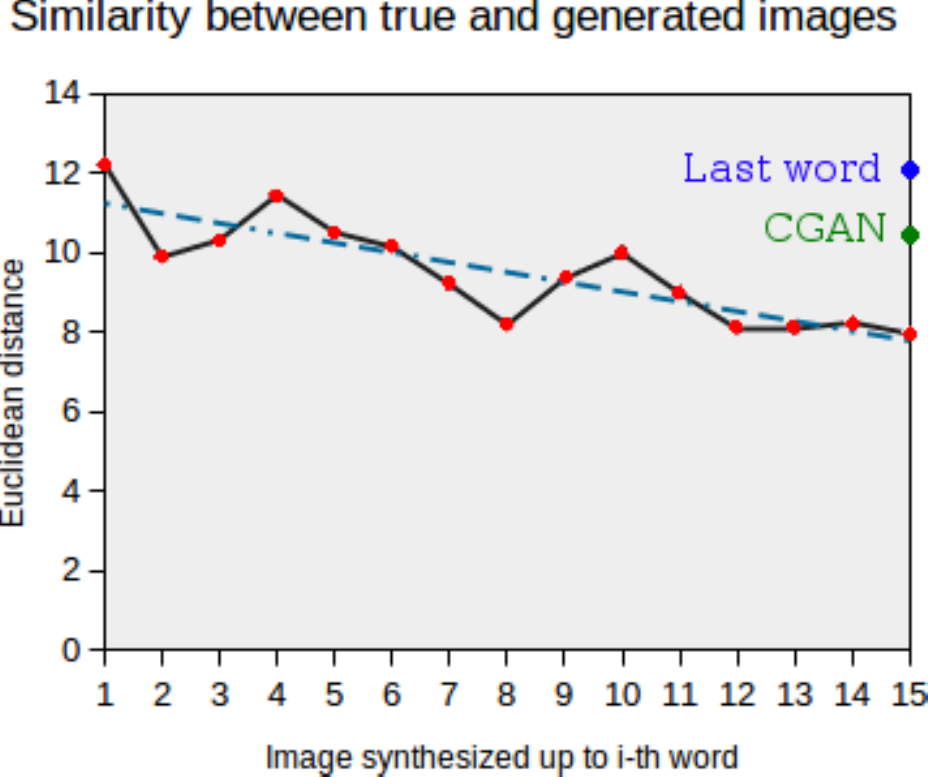}}}
  &
  \resizebox{0.245\textwidth}{!}{\rotatebox{0}{
  \includegraphics{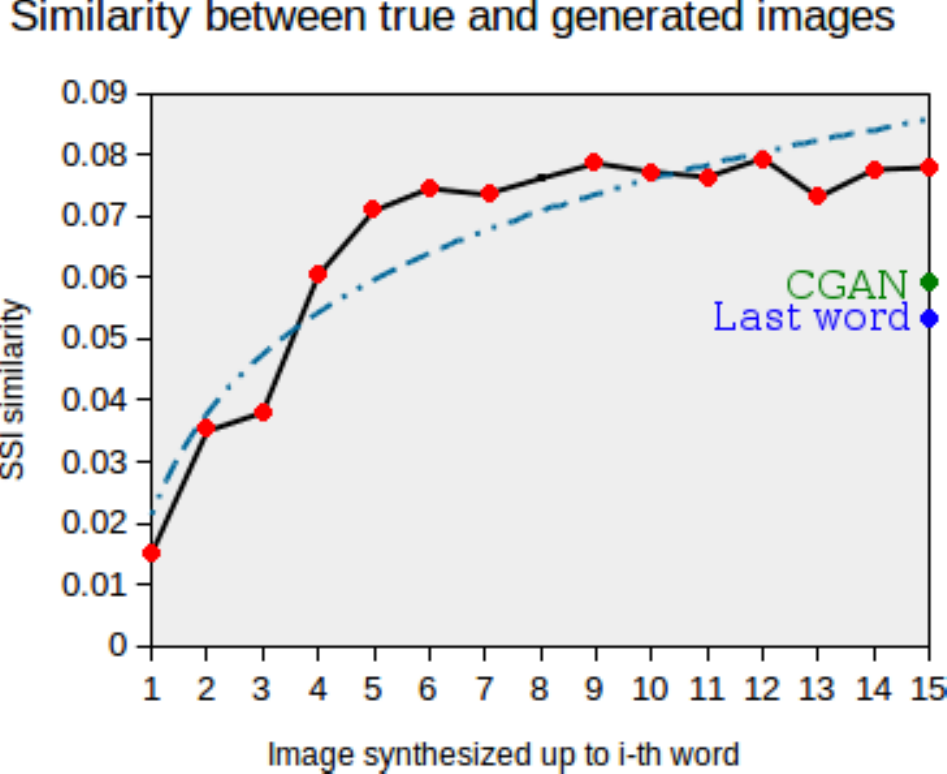}}}
  \\
  Bird VGG-16 & Bird SSI
  \\
  \resizebox{0.24\textwidth}{!}{\rotatebox{0}{
  \includegraphics{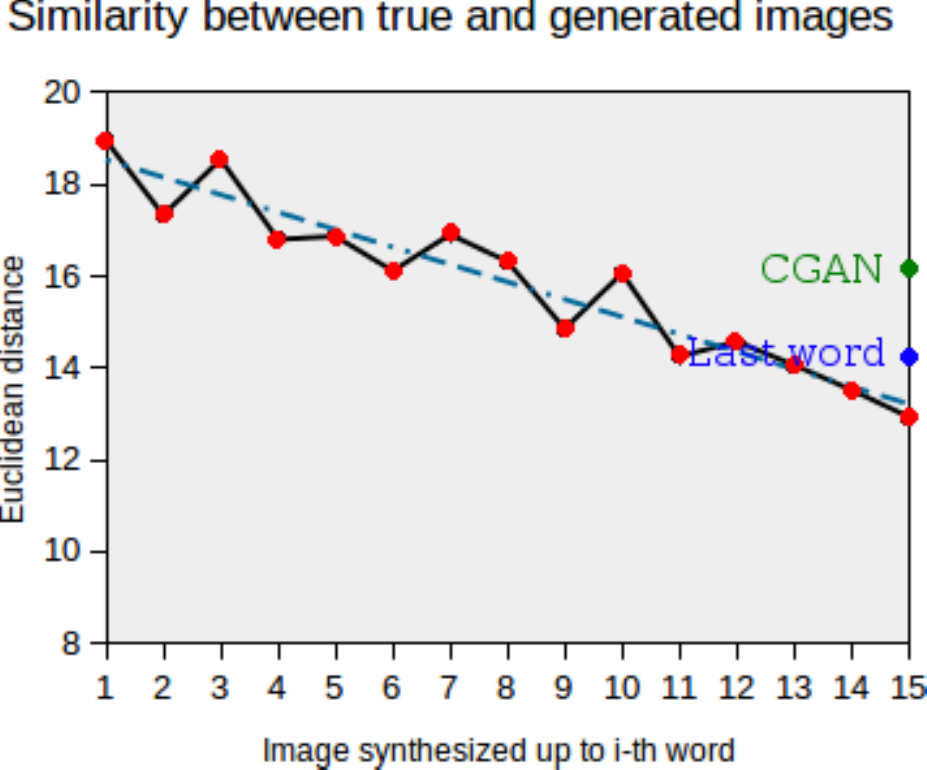}}}
  &
  \resizebox{0.245\textwidth}{!}{\rotatebox{0}{
  \includegraphics{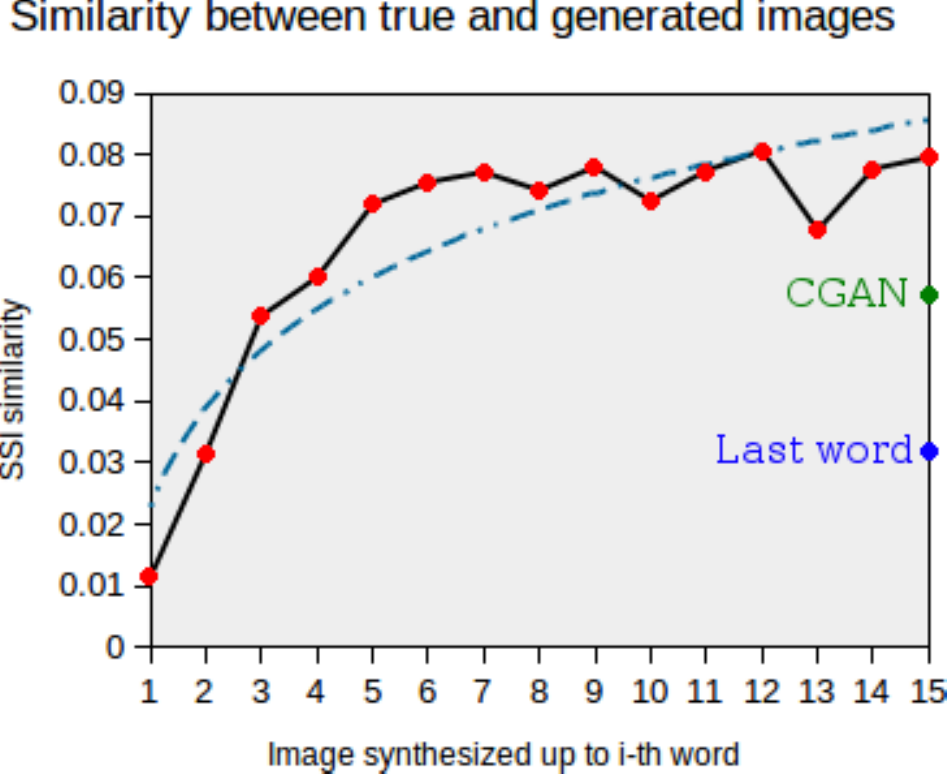}}}
  \\
  Flower VGG-16 & Flower SSI
  \\
\end{tabular}}
\caption{SSI \cite{wang2004image} and Euclidean distance between real images and generated images calculated using features extracted from VGG-16 \cite{Simonyan14c}. Results are compared to two methods which are \cite{reed2016generative} (CGAN) and generating images using only the output of the last step of the proposed LSTM conditional GAN (Last-word). To show the proposed approach can gradually achieve better results than the compared two methods, we measure the similarity for images generated using the proposed approach up to each word in sentences and show the trend using dashed lines in the figures.   }
\label{fig: quantitative results}
\end{figure} 

Since the conditional GAN generates only one image for each sentence, to compare it to the proposed network, we choose the image generated at the last word of a sentence in the comparison. In Figure~\ref{fig: cgan comparison}, we can see that our method get the most realistic and clearest results which also have the best correspondence.

\subsection{Quantitative comparison}  
We evaluated the proposed network quantitatively by measuring the similarity between synthesized images and corresponding real images. We measured the similarity using two different methods. First, we used features extracted from VGG-16 \cite{Simonyan14c} layer as representations of images and compare image similarities based on these features. Without loss of generality, we combined features from the second layer and the second last layer to cover both general and detailed feature of the images. Then, similarity could be calculated as a Euclidean distance between features. Second, we measured Structural Similarity Index (SSI) between generated images and corresponding real images. SSI incorporates luminance and contrast masking into account. Strong inter-dependencies of closer pixels are also included in the error calculation, and metric is computed on small windows of the images. The value of SSI is between -1 and 1 (higher is better). We showed results of similarity measured in Figure \ref{fig: quantitative results}. 

The proposed LSTM conditional GAN generates images at each word in a sentence, and this is the primary contribution of this work. We claim that the proposed approach can generate images with better quality by doing so. To validate our claim, we measured the similarity between real images and images generated by the proposed approach at each word of a sentence. We compare results to another two methods \cite{reed2016generative} (CGAN) and generating images using the last step of the proposed LSTM conditional GAN (Last-word). From Figure \ref{fig: quantitative results}, it could be seen that both VGG-16 and SSI similarity measures validate our claim and results of the proposed approach is clearly better than the other two compared methods. Both VGG-16 and SSI measures could conclude that the proposed network can generate images which can capture the content of real images both in small and large scales. And this is primarily due to the ability of generating images at each word of description, which explore more semantics and hidden information of descriptions.


\section{Conclusion}
In this paper, we built up a new neural network architecture to generate images of real-life objects. We demonstrated that our model could generate more realistic images given the description of Oxford-102 Flowers dataset and Caltech-UCSD Bird-200-2011 datasets. We also showed the capability of our network that visualizes image results for a sentence word by word. At last, we demonstrated that our proposed model gets the best quantitative results than other methods on Oxford-102 Flowers dataset. In future work, we hope to get more realistic images from descriptions.






%
{\small
\bibliographystyle{ieee}
\bibliography{egbib}
}

\end{document}